\documentclass[sigconf]{acmart}

\AtBeginDocument{%
  \providecommand\BibTeX{{%
    \normalfont B\kern-0.5em{\scshape i\kern-0.25em b}\kern-0.8em\TeX}}}

\setcopyright{acmcopyright}
\copyrightyear{2018}
\acmYear{2018}
\acmDOI{10.1145/1122445.1122456}

\acmConference[Woodstock '18]{Woodstock '18: ACM Symposium on Neural
  Gaze Detection}{June 03--05, 2018}{Woodstock, NY}
\acmBooktitle{Woodstock '18: ACM Symposium on Neural Gaze Detection,
  June 03--05, 2018, Woodstock, NY}
\acmPrice{15.00}
\acmISBN{978-1-4503-XXXX-X/18/06}

\begin{document}

\title{Beyond Point Estimate: Inferring Ensemble Prediction Variation from 
Neuron Activation Strength in Recommender Systems}

%
\author{Zhe Chen, Yuyan Wang, Dong Lin, Derek Zhiyuan Cheng, Lichan Hong, Ed H. Chi, Claire Cui}
\affiliation{Google, Inc.\\
\{chenzhe,yuyanw,dongl,zcheng,lichan,edchi,claire\}@google.com}
\email{}
\email{}
\email{}


\newcommand{\eg}{{e.g.}}
\newcommand{\Eg}{{E.g.}}
\newcommand{\ie}{{i.e.}}
\newcommand{\Ie}{{I.e.}}

\newcommand{\movielens}{{MovieLens}}
\newcommand{\mr}{{MovieLens-R}}
\newcommand{\mc}{{MovieLens-C}}
\newcommand{\criteo}{{Criteo}}

\newcommand{\uncertainty}{{prediction variance}}

\newcommand*\shirley[1]{\color{red}{\textbf{#1}}}
\newcommand*\todo[1]{\color{red}{\textbf{#1}}}
\newcommand*\dongl[1]{\color{blue}{\textbf{#1}}}
\newcommand*\yuyanw[1]{\color{purple}{\textbf{#1}}}

\begin{abstract}

Despite deep neural network (DNN)'s impressive prediction performance in various domains,
it is well known now that a set of DNN models trained with the same model
specification and the same data can produce very different prediction results. 
Ensemble method is one state-of-the-art benchmark for prediction uncertainty estimation. 
However, ensembles are expensive to train and serve for web-scale traffic. 

In this paper, 
we seek to advance the understanding of prediction variation estimated by the ensemble method. 
Through empirical experiments on two widely used benchmark datasets Movielens and Criteo in recommender systems, 
we observe that prediction variations come from various randomness sources, 
including training data shuffling, and parameter random initialization. 
By introducing more randomness into model training, we notice that ensemble's mean predictions tend to be more accurate while the
prediction variations tend to be higher. 
Moreover, we propose to infer prediction variation from neuron activation strength and demonstrate the strong prediction power from activation strength features. 
Our experiment results show that 
the average $R^2$ on \movielens\ is as high as 0.56 
and on \criteo\ is 0.81.
Our method performs especially well when detecting the lowest and 
highest variation buckets, with 0.92 AUC and 0.89 AUC respectively. 
Our approach provides a simple way for prediction variation estimation, which opens up new opportunities for future work in many
interesting areas (\eg, model-based reinforcement learning) 
without relying on serving expensive ensemble models.

\end{abstract}

%

\settopmatter{printacmref=false} 
\renewcommand\footnotetextcopyrightpermission[1]{} 
\pagestyle{fancy}
	\lhead{\footnotesize{Beyond Point Estimate: Inferring Ensemble Prediction Variation from 
Neuron Activation Strength in Recommender Systems}}
    \rhead{\footnotesize{arxiv or preprint later}}

\maketitle

\section{Introduction}

Deep neural networks (DNNs) have gained widespread adoption in recent years across many domains.
Despite their impressive performance in various applications, 
most DNNs today only generate point predictions.
And it is well known that a set of DNN models trained with the same model 
specification and the same data can produce very different 
predictions~\cite{lakshminarayanan2017simple, ovadia2019can, wen2020batchens}. 
Researchers realize that point predictions do not 
tell the whole story and raise questions about whether DNNs 
predictions can be trusted~\cite{jiang2018trust, schulam2019can}. 

In response, a growing number of researches are looking into quantifying 
prediction uncertainties for DNNs. 
Ensemble method is a state-of-the-art benchmark for prediction 
uncertainty estimation to consolidate agreements 
among the ensemble members and produce better point predictions~\cite{breiman1996bagging, dietterich2000ensemble, lakshminarayanan2017simple, ovadia2019can}.
Many researches focus on 
whether these point predictions are well-calibrated on either 
in-distribution or out-of-distribution (OOD) 
data~\cite{ovadia2019can, guo2017calibration, lee2017training,liu2020simple}. 
However, there exist disagreements among the predictions 
 in the ensemble, which we call {\bf \em prediction variation}. 
For example, different models in the ensemble often 
yield different prediction results even on the same input example. 
Although prediction variation contributes to model 
uncertainty~\cite{lakshminarayanan2017simple}, 
there is no comprehensive study to advance the understanding of it. 

Ensembles provide us with a good approximation of 
prediction variation, but they are computationally expensive as they 
require training multiple copies of the same model. 
At inference time, they require computing predictions on every 
example for every ensemble member, which can be infeasible for real-time large-scale machine learning systems. 
Researches propose various resource-saving 
techniques \cite{gal2016dropout, huang2017snapshot, wen2020batchens, lu2020uncertainty}. 
However, as far as we know, none of these works studies whether we can 
infer prediction variation from neuron activation 
strength collected from the DNN directly,
without running predictions on the same data multiple times. 
Here, we use {\bf \em neuron activation strength} to indicate DNN's neuron output 
strength, \eg, the value of the neuron after activation and whether the neuron is activated.

We hypothesize that neuron activation strength could 
be directly used to infer prediction variation. 
Our intuition is based on neuroscience's
Long-Term Potentiation (LTP) process~\cite{nicoll2017brief} 
which states that connections between neurons become stronger 
with more frequent activation. 
LTP is considered as one of the underlying mechanisms for learning and 
memorization. 
If we imagine deep networks learn like the brain, 
some groups of neurons will be more frequently and/or strongly activated, i.e. 
strengthened neurons. During learning, 
these strengthened neurons represent where the network has learned or memorized better. 
Therefore, we hypothesize that deep neural networks predict more 
confidently if an input activates through strengthened neurons, 
and less so if the input goes through weaker neurons.

\vspace{0.1cm}
\noindent 
{\bf Our Goal} --- 
In this paper, we aim to advance the understanding of 
prediction variations 
estimated by ensemble models, and look into 
the predictive power of neuron activation strength on 
prediction variations.  
To the best of our knowledge, we are the first to 
conduct comprehensive studies on prediction variations from different ensemble models, 
and we are also the first to demonstrate that we are able to infer  
prediction variation from neuron activation strength. 

\vspace{0.1cm}
\noindent 
{\bf Challenges} --- We face the following challenges: 

{\em Variation Quantification} ---
There are a variety of prediction problems. For example, 
predicting the target rating for a user on a given movie could be a regression task
or a multi-class 
classification task by dividing the movie ratings into multiple buckets; 
predicting click-through rate is often modeled as a binary classification task. 
There is no standard way
to quantify prediction variation for such a variety of
tasks.

{\em Variation Sources} --- 
Training a set of models with the same model specification and
the same data can produce very different results, and 
the prediction disagreements are inherently caused by the nonconvex 
nature of DNN models in which multiple local minima exist. 
Multiple randomness sources could lead to the disagreements, such as 
random initialization of DNN parameters, random shuffling of training data,
sub-sampling of training data, optimization algorithms, and even the hardware itself. 
It is often hard to identify the contribution of each randomness source 
to prediction variation. 

{\em Neuron Activation Strength} --- 
We hypothesize that a deep network's 
prediction variations are strongly correlated with neuron activation 
strength throughout training and at inference time. 
However, it is not straightforward to demonstrate this relationship for 
different prediction problems and randomness sources.

\vspace{0.1cm}
\noindent 
{\bf Our Approach} --- 
In this paper, we investigate sources for prediction variation, and by 
controlling the randomness sources explicitly, we
demonstrate that neuron activation strength has strong 
prediction power to infer ensemble prediction variations
in almost all the randomness-controlled settings.
We demonstrate our findings on two popular datasets used to
evaluate recommender systems, MovieLens and Criteo.

First, we quantify prediction variation 
across different types of target tasks, including
regression, binary classification, and multi-class classification tasks. 
We use standard deviation of the ensemble predictions
to quantify prediction variation for regression 
and binary classification tasks, and use 
KL divergence based measurement to quantify prediction distribution 
disagreement for multi-class classification tasks. 

Second, we identify and examine three variation sources of randomness
(data shuffling, weight initialization, and data re-sampling). By
explicitly controlling randomness sources, 
we study their contributions to the performance of 
point prediction and prediction variation. 
Our results show that every variation source 
exhibits a non-negligible contribution towards the 
total prediction variation. 
The prediction variation 
may have different sensitivities to different types of 
variation sources on different target tasks.  
When we include more variation sources, 
the ensemble's prediction mean tends to be more accurate, while the 
prediction variations are higher.

Finally, we demonstrate that neuron activation strength 
has strong prediction power in estimating prediction variation of 
ensemble models, while the neuron activation strength information 
can be obtained from a single DNN. We obtain the neuron activation strength information 
from the neural network. With this strength information, we can add a cheap auxiliary 
task to estimate prediction variation directly. 
Our experiment results show that
our activation strength based method estimates prediction variation fairly well 
as a regression task.  
The average $R^2$ on \movielens\ is 0.43 and 0.51 with
different task definitions, 
and on \criteo\ is 0.78. 
Our method is especially good at detecting the lowest and highest variation 
bucket examples, on average with 0.92 AUC score for the lowest bucket and 0.89 AUC score
for the highest bucket on both datasets. 
Our approach is complementary and orthogonal to 
many other resource-saving or single model prediction
variation estimation techniques, as it doesn't alter the target task's 
optimization objectives and process.  

\vspace{0.1cm}
\noindent 
{\bf Applications} --- 
Prediction variation quantification is 
a fundamental problem and 
our activation strength based approach opens up new opportunities 
for a lot of interesting applications. 
For example, 
in model-based reinforcement learning, prediction variation 
has to be quantified for exploration~\cite{zhou2019neural,chua2018deep}. 
In curriculum learning~\cite{bengio2009curriculum}, 
prediction variation can be used 
as a way for estimating example difficulty. 
In medical domain, prediction variation can be used to capture 
significant variability in patient-specific predictions~\cite{dusenberry2020analyzing}. 
Our proposed activation strength based method provides 
a simple and principled way to serve prediction variation 
estimate by deploying a cheap auxiliary task, 
instead of using an expensive ensemble model, during inference time.
We explore applying our method in 
the above scenarios.

\vspace{0.1cm}
\noindent {\bf Contributions} --- Our contributions are four fold: 
\begin{itemize}
    \item Framework for prediction variation estimation using activation 
    strength in Section~\ref{sec:preliminary}.
    \item Formal quantification on prediction variation for various 
    target tasks in Section~\ref{sec:variation_definition}. 
    \item Prediction variation understanding by explicitly controlling
    various randomness sources in Section~\ref{sec:prediction_sources}. 
    \item Empirical experiments to demonstrate strong predictive 
    power from neuron activation strength to estimate ensemble prediction variation 
    in Section~\ref{sec:variation_estimation}. 
\end{itemize}

We cover the related work in Section~\ref{sec:related_work}, 
and conclude with a discussion of future work in Section~\ref{sec:conclusion}.

\vspace{-0.2cm}
\section{Related Work}
\label{sec:related_work}

In machine learning literature, researchers mostly 
focus on two distinct types of uncertainties: 
aleatoric uncertainty and epistemic uncertainty~\cite{der2009aleatory}. 
Aleatoric uncertainty is due to the
stochastic variability inherent in the data generating 
process~\cite{liu2019accurate}. 
Aleatoric uncertainty corresponds to  
{\em data uncertainty}, which describes uncertainty for a given
outcome due to incomplete information~\cite{knight1921risk}. 
Epistemic uncertainty is due to our lack of knowledge about the data generating mechanism~\cite{liu2019accurate}, and 
corresponds to {\em model uncertainty}, which 
can be viewed as uncertainty regarding the
true function underlying the observed 
process~\cite{bishop2006pattern}. 
In this paper, we focus on studying model uncertainty, 
especially prediction variations or disagreements.

There has been extensive research on methodologies for estimating model uncertainty and discussions on their comparisons~\cite{ovadia2019can}. Principled approaches include Bayesian 
approaches~\cite{neal2012bayesian,mackay1992practical,hinton1993keeping,louizos2017multiplicative,zhu2018bayesian} and ensemble-based approaches~\cite{lakshminarayanan2017simple}. 
Bayesian methods provide a mathematically grounded framework to model uncertainty, through learning the deep neural network as Gaussian processes~\cite{neal2012bayesian}, or learning approximate posterior distributions for all or some weights of the network ~\cite{blundell2015weight,kwon2018uncertainty}. Ensemble methods~\cite{lakshminarayanan2017simple}, on the other hand, is a conceptually simpler way to estimate model uncertainty. There are multiple ways to create ensembles of neural networks: bagging~\cite{breiman1996bagging}, Jackknife~\cite{mcintosh2016jackknife}, random parameter initialization, or random shuffling of training examples. The resulting ensemble of neural networks contains some diversity, and the variation of their predictions can be used as an estimate of model uncertainty. 
A lot of research work results are based on the ensemble 
method~\cite{berthelot2019mixmatch, de2018clinically, chua2018deep, leibig2017leveraging,ovadia2019can}. 
For example, 
\cite{de2018clinically} demonstrates promising results 
uses deep ensembles for diagnosis and referral in retinal disease. 
\cite{chua2018deep} proposes a new algorithm for 
model-based reinforcement learning 
by incorporating uncertainty via ensemble.  
In this paper, we use ensemble as the ground truth to produce prediction variations 
in different scenarios (\ie, different randomness settings). 

Estimating model uncertainty through Bayesian modeling or ensemble usually incurs significant computation cost. For example, Bayesian neural networks that perform variational learning on the full network~\cite{blundell2015weight} significantly increase the training and serving cost. The cost for ensemble methods scales by the number of models in the ensemble, which can be prohibitive for practical use. To this end, researchers have proposed various techniques to reduce the cost for Bayesian modelling and ensemble methods. For example, single-model approaches are proposed to quantify model uncertainty by modifying the output layer~\cite{tagasovska2019single, liu2020simple}, deriving tractable posteriors from last layer output only~\cite{riquelme2018deep, snoek2015scalable}, or constructing pseudo-ensembles that can be solved and estimated analytically~\cite{lu2020uncertainty}. Our proposed method in this paper can also be viewed as a single-model approach for model uncertainty estimation. However, we do not impose any Bayesian assumptions on the network or any distributional assumptions on the ensembles. Instead, we build an empirical model to learn the association between activation strength and model uncertainty, and use it to estimate model uncertainty for new examples, which offers a relatively simple, robust and computationally efficient way to estimate prediction variation from a single model.

\section{Variation Estimation Framework}
\label{sec:preliminary} 

\begin{figure}[t!]
    \includegraphics[width=0.9\linewidth]{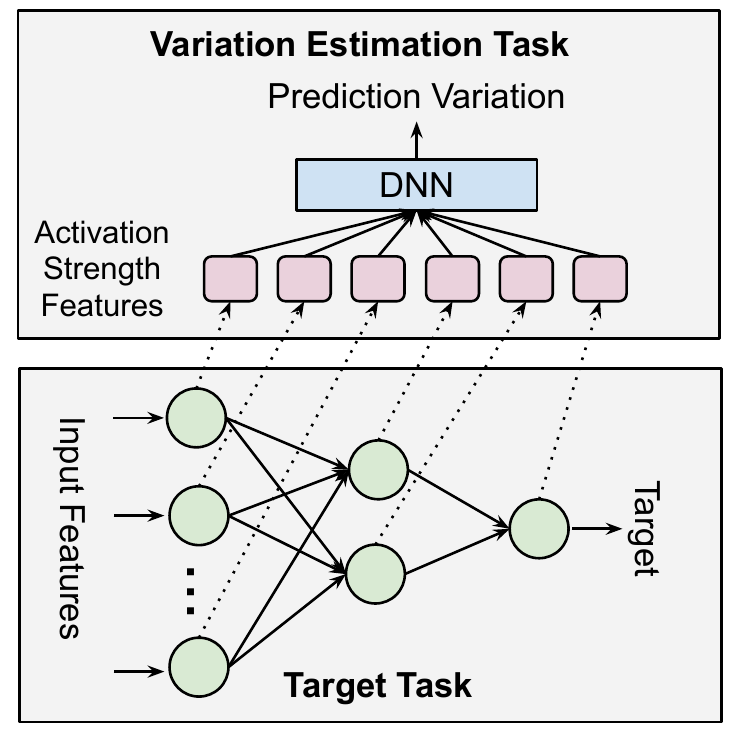}
    \caption{Our framework for prediction variation 
    estimation using activation strength.}
    \label{fig:framework}
\end{figure}

Similar to the work on model uncertainty 
estimation~\cite{nix1994estimating,nix1995learning,su2018tight}, 
we build two components for the prediction variation estimation framework: 
target task, and variation estimation task, 
as shown in Figure~\ref{fig:framework}. Before discussing 
the two tasks in detail, 
we first introduce the two experiment datasets that 
we use throughout this paper. 

\subsection{Datasets}
\label{sec:datasets}

Our studies are based on two datasets: \movielens\ and \criteo.

\vspace{0.1cm}
\noindent {\bf \movielens} ---
The \movielens~\footnote{\url{http://files.grouplens.org/datasets/movielens/ml-1m-README.txt}} contains 1M movie ratings from 6000 users on 4000 
movies. This data also contains 
user related features and movie related features.

\vspace{0.1cm}
\noindent {\bf \criteo} --- 
The Criteo Display Advertising challenge
~\footnote{https://www.kaggle.com/c/criteo-display-ad-challenge} 
features a binary classification task to predict click-through rate
(clicked event's label is 1, otherwise 0). 
The \criteo\ data consists of around 40M examples 
with 13 numerical and 26 categorical features.

\subsection{Target Task}
\label{sec:target_tasks}

The target task is defined by the original 
prediction problem, 
such as the rating prediction task on \movielens, 
and the click-through prediction task on \criteo. 
The target task
takes in the input features from the dataset,
and predicts the target.  
In this paper, we focus on the multi-layer perceptron 
architecture (MLP), 
with ReLU as the activation function for all layers. 
Furthermore, we define three target tasks 
on \movielens\ and \criteo. 

\vspace{0.1cm}
\noindent {\bf \movielens\ Regression (\movielens-R)} --- 
    The target task takes in 
    user-related features (\ie, id, gender, age, and occupation) 
    and movie-related features (\ie, id, title and genres), and 
    predicts movie rating as a regression task. 
    The movie ratings are integers from 1 to 5.
    We use mean squared error (MSE) as the loss function. 
    MSE is a standard metric for evaluating the 
    performance of rating prediction in 
    recommenders~\cite{herlocker2004evaluating,saadati2019movie,bennett2007netflix}. 
    For example, \cite{bennett2007netflix} used 100 million 
    anonymous movie ratings
    and reported their Root Mean Squared Error (RMSE)
    performance on a test dataset as 0.95.

Each model trains for 20 epochs with early-stopping. 
We only use the observed ratings in \movielens\ as training data. 
Similar to the neural collaborative filtering 
framework~\cite{he2017neural}, 
we use fully connected neural network for the rating 
prediction task and ReLU as the activation function.
We set the fully connected neuron layer sizes to be 
[50, 20, 10]. 
We set the user id and item id embedding size to 8~\cite{he2017neural}, 
the user age embedding size to 3, and user 
occupation embedding size to 5. 

\vspace{0.1cm}  
\noindent {\bf \movielens\ Classification (\movielens-C)} --- 
    Similar to the \movielens\ regression task, we predict
    movie ratings as 5 integer values from 1 to 5 and 
    model this problem as a multi-class classification 
    task with softmax cross entropy as the loss function.

We experiment with temperature scaling
values 
$T=$[0.1, 0.2, 0.5, 1, 2, 5, 10] with batch size of 1024 to make sure predictions are well 
calibrated~\cite{guo2017calibration}. We pick $T=0.2$ which gives the best Brier score\footnote{
\url{https://en.wikipedia.org/wiki/Brier_score}} while 
achieving similar accuracy compared to other settings. 
    
\vspace{0.1cm}    
\noindent {\bf \criteo} --- This target task uses
    a set of numerical and categorical features to predict the click-through rate. 
    The label for the task is either 0 or 1 representing whether an ad is clicked or not. 
    We model this problem as a binary classification 
    task with sigmoid cross entropy loss function. 
    The trained model outputs a float between 0 and 1 
    representing the predicted click-through probability.

We use the same model setting as described in~\cite{ovadia2019can}, except 
for ReLU layer sizes. In the beginning, we set ReLU layer sizes to 
[2572, 1454, 1596] as in~\cite{ovadia2019can}, 
but found that only around 80 neurons are activated at least once on a 10k sample data. 
As a result, for the experiments in this paper, we use ReLU layer sizes of [50, 20, 10] and 
find the prediction performance is similar to the model with much larger ReLU layer sizes. Each model is trained for 1 epoch.

\subsection{Variation Estimation Task}

In this paper, we focus on using neuron activation strength to estimate 
prediction variation for each input example. 
We use ensemble to estimate prediction variation as the 
ground-truth label.  
We define prediction variation 
formally in Section~\ref{sec:variation_definition}. 

As shown in 
Figure~\ref{fig:framework}, 
we build a neural network model taking the 
 neuron activation strength features to estimate 
 prediction variation. 
During the inference time, we
directly output the estimated  
prediction variation using activation strength 
as an auxiliary task. 
In our current setup, we collect the activation strength features 
from all the neurons in the target task. Also, we find 
that it is possible to identify important neurons 
in order to reduce the number of activation strength features. 
Due to the space limit, we will not discuss feature reduction
in this paper. 
The detailed setup of the variation estimation task is discussed in 
Section~\ref{sec:variation_estimation}.

\section{Prediction Variation Quantification}
\label{sec:variation_definition}

In this section, we formally quantify prediction 
variation in different problem settings.
Ensemble is one state-of-the-art benchmark for prediction 
uncertainty estimation~\cite{breiman1996bagging, dietterich2000ensemble, lakshminarayanan2017simple, ovadia2019can}. 
We use ensemble to estimate model prediction variation, that is how much 
disagreement there is among ensemble model predictions.

Given the same training data and model configuration, 
we train an ensemble of $N$ models $M = \{m\}$, where 
$M$ is the set of models, and $N$ is the ensemble size.
Let $\{x\}$ be the testing data, and 
$x$ represents the feature vector. 
Each of the trained model $m \in M$ makes a prediction on an example 
$x \in \{x\}$ as $y^\prime_{m}(x)$.

For {\em regression and binary classification} tasks, the model output is a float 
value, thus we define prediction variation as 
{\em value prediction variation} based on the standard deviation 
of the predicted float values across different models 
in the ensemble. 
For {\em multi-class classification} tasks, the model output is
a probability distribution over different categories. 
Thus, we define prediction variation
as {\em distribution prediction variation} based on the 
KL-disagreement or generalized Jensen-Shannon 
divergence~\cite{lakshminarayanan2017simple} 
on the predicted probability distributions. 
Now we define prediction variation formally.

\begin{definition}{(Value Prediction Variation)}
\label{def:regression_uncertainty}
Given an example $x$ that represents the feature vector, 
we define its prediction variation $PV(x)$ to be 
the standard deviation of predictions from the set of models $M$ as 
$PV(x) = \sqrt{\frac{\sum_{m \in M}{(y^\prime_{m}(x) - \bar{y}(x))^2}}{|M|-1}}$, 
where $\bar{y}(x) = \frac{1}{|M|} \sum_{m \in M} y^\prime_{m}(x)$
\end{definition}

\begin{definition}{(Distribution Prediction Variation~\cite{lakshminarayanan2017simple})}
\label{def:classification_uncertainty}
Given the example $x$ that represents the feature vector, 
let the prediction distribution for the example $x$ be $p(y|x)$. 
We define prediction variation $PV(x)$ 
to be the sum of the Kullback-Leibler (KL) divergence 
from the prediction distribution of each model $m \in M$ 
to the mean prediction distribution of the ensemble. 
$ PV(x) = \sum_{m=1}^{M} KL(p_m(y|x) || p_E(y|x))$ where $p_E(y|x) = M^{-1} \sum_{m} p(y|x)$ is the mean prediction of the ensemble.
\end{definition}

\begin{table*}[t]
\begin{tabular}{l|ccccc|ccc|cccc}
\toprule
Randomness    & \multicolumn{5}{c|}{MovieLens-R} & \multicolumn{3}{c|}{MovieLens-C} &  \multicolumn{3}{c}{Criteo} \\ \cline{2-13}
Settings     & MSE  &ACC & PV Mean  & PV Std & PV Coeff
     &ACC & PV Mean & PV Std  
     & AUC  & PV Mean  & PV Std & PV Coeff\\ \hline
(R0) None  & 0.7980  & 0.4483 & 0.0000  & 0.0000  & 0.00\%
& 0.4635  & 0.0000  & 0.0000  
&0.7956 & 0.0000  & 0.0000  &0.00\%\\ 
(R1) R  
& 0.7569  &0.4570 & 0.1948  & 0.0692 &5.83\%
& 0.4818 & 4.4486 & 2.5744
&0.7991 & 0.0300 & 0.0162 &16.3\% \\
(R2) S
& 0.7671  &0.4473 & 0.1433  & 0.0509 & 4.36\%
& 0.4746 & 2.7379 & 1.6717
&0.7999 & 0.0359 & 0.0179 &20.7\%\\
(R3) R+S
& 0.7479 &0.4521 & 0.1936  & 0.0649 & 5.87\%
& 0.4829 & 4.4637 & 2.5053
&0.7999 & 0.0358 & 0.0181 &20.4\%\\ \hline
(R4) J
& 0.7718 &0.4464 & 0.1494   & 0.0638 & 4.54\%
& 0.4745 & 2.7522 & 1.9771
&{\bf 0.8003} & 0.0394 & 0.0198 &22.9\%\\
(R5) R+J
& 0.7489 &0.4522  & {\bf 0.2035}  & {\bf 0.0701} & {\bf 6.14\%}
& 0.4829  & 4.7250 & 2.6514
&0.8002 & 0.0396 & 0.0199 &22.9\%\\
(R6) S+J
& 0.7640 &0.4486  & 0.1560 & 0.0571 & 4.74\%
& 0.4766  & 3.1739 & 1.9803
&0.8002 & {\bf 0.0409} & 0.0200 &23.5\% \\
(R7) R+S+J
& {\bf 0.7457} & {\bf 0.4528}  & 0.2013 & 0.0667 & 6.08\%
& {\bf 0.4838}  & {\bf 4.7332} & {\bf 2.6723}
&0.8000 & 0.0407 & {\bf 0.0202} & {\bf 23.6\%}
\\
\bottomrule
\end{tabular}
\caption{Ensemble's prediction 
accuracy and prediction variation (PV) on 8 randomness settings.
PV coefficient (coeff) shows the average ratio of PV to the ensemble mean prediction over all the testing examples. }
\label{tab:randomness_control}
\end{table*}

\begin{figure*}[t!]
    \centering
    \includegraphics[width=.33\linewidth]{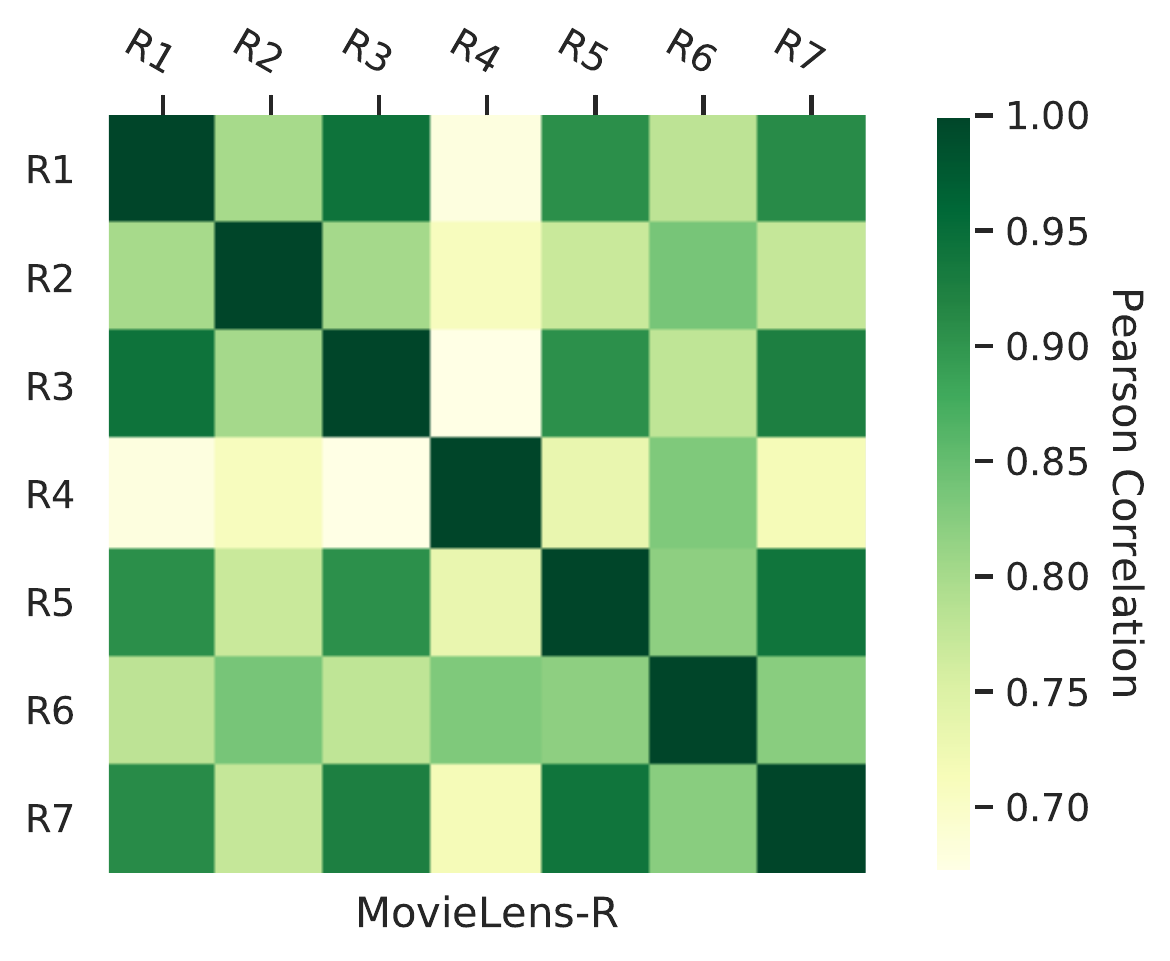}
    \includegraphics[width=.33\linewidth]{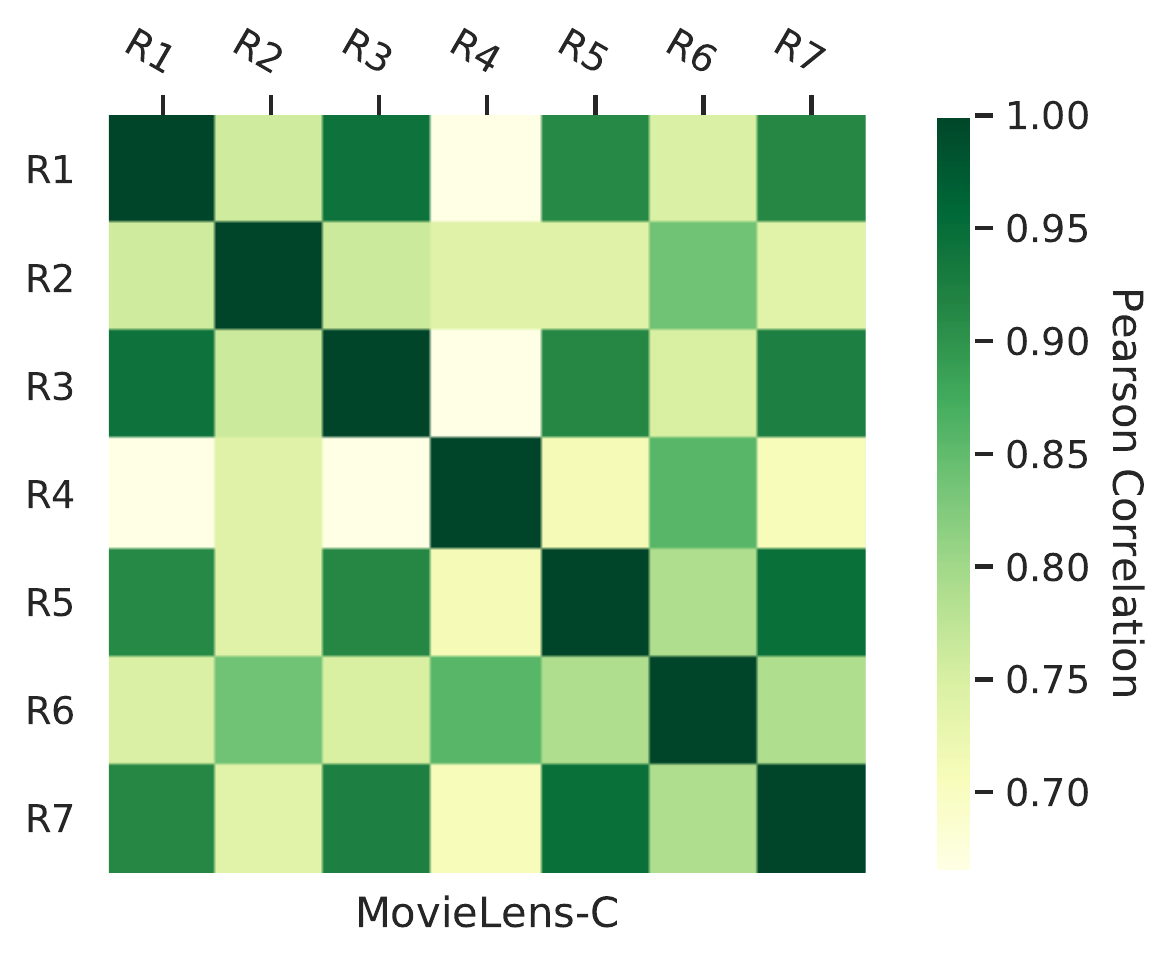}
    \includegraphics[width=.33\linewidth]{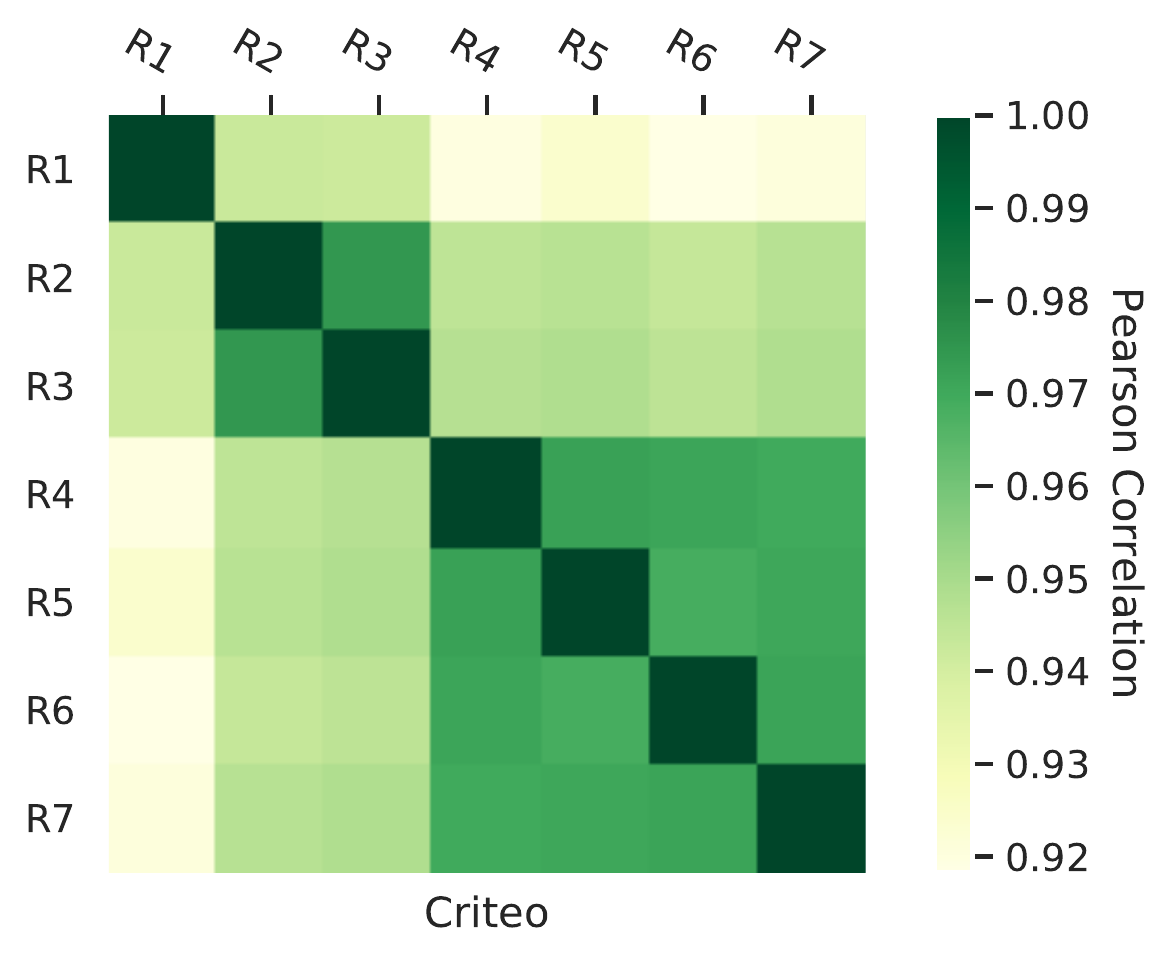}
    \vspace{-0.4cm}
    \caption{Pearson correlation of prediction variations for ensembles with different randomness settings.}
    \label{fig:randomness_correlation}
\end{figure*}

\section{Prediction Variation Sources}
\label{sec:prediction_sources}

In this section, we diagnose the prediction variation sources, and 
we are interested in seeing their effects on the total prediction variation by
controlling type of each randomness source. 

There are many sources contributing to prediction variation.
Random initialization of DNN parameters contributes randomness to model predictions. 
Randomness can also come from training data shuffling 
or sub-sampling. 
In addition, asynchronous or distributed training could lead to training order randomness. 
More surprisingly, we observe the hardware itself contributes to the model prediction
variation: we find that by fixing all the other settings, training a model 
on different CPUs might produce different models.

In this paper, we consider three types of randomness sources:

\vspace{0.1cm}
\noindent {\bf 1. Shuffle (S)} --- Whether randomly shuffles 
input data, \ie, randomizes input data order. 

\vspace{0.1cm}
\noindent {\bf 2. RandInit (R)} --- 
Whether randomly initializes model parameters, 
including DNN weights and embeddings. 
We can fix the initialization by setting 
a global random seed in Tensorflow.

\vspace{0.1cm}
\noindent {\bf 3. Jackknife (J)} --- 
Whether randomly sampling input data by 
applying delete-1 
Jackknife~\cite{mcintosh2016jackknife}.
We split data into N Jackknife sub-samples and 
each ensemble member randomly leaves one Jackknife 
sample out. 
We use 100 models for each ensemble as discussed in Appendix~\ref{sec:ensemble_size}, 
and we split the data into 100 unique Jackknife sub-samples. 
Another popular data sampling method is bootstrap~\cite{wichmann2001psychometric}. 
For this work, we pick Jackknife due to its simplicity 
to implement.

\vspace{0.1cm}

We set up the prediction variation randomness control experiments as follows: 
First, if incorporating Jackknife randomness, we obtain the delete-1 Jackknife 
sub-samples; otherwise, we use all the training data. 
Second, if incorporating Shuffle randomness,
we shuffle the training data; otherwise we do not.  
Finally, if incorporating RandInit randomness, 
we randomly initialize all the parameters without a fixed global seed 
for all the ensemble members; otherwise we use a fixed global seed.\footnote{When RandInit is enabled, we use a fixed set of 100 random seeds.}
We use an ensemble of 100 models for each of the randomness settings:
as discussed in Appendix~\ref{sec:ensemble_size}, 
93\% of prediction variation in the 
ensemble of 1000 models can be captured with size 100 ensemble. 

On each dataset of \movielens\ and \criteo, we randomly split the data into 
training and testing. 
On \movielens\ we split the 1M data into 60\% for training, 
and 40\% for testing. On \criteo, 
same to~\cite{ovadia2019can}, we use 37M data for training 
and 4.4M for testing.
We obtain the prediction variation estimation 
on all the testing examples for further analysis.

\vspace{0.1cm}
\noindent {\bf Randomness Source Comparison} ---
Table~\ref{tab:randomness_control} shows the 
accuracy and prediction variation statistics
for different combinations of the three randomness sources. 
For each type of randomness combination (\eg, (R3) R+S 
means using RandInit and Shuffle only),  
we train 100 models of the same setting, and 
obtain the ensemble mean prediction for accuracy evaluation
and report the prediction variations. 
For accuracy evaluation, we report 
Mean Squared Error (MSE) and accuracy (ACC) for \mr, 
ACC for \mc, and AUC score for \criteo. 
We obtain ACC for \mr\ by rounding the ratings to the closest integers. 
For prediction variation metrics, 
we report prediction variation (PV) mean and standard deviation, 
and PV coefficient (coeff). 
We obtain the PV coefficient for each example 
$x$ as PV(x) divided by the ensemble mean prediction. 

From the tables, we can see that 
each type of the randomness sources exhibits a
non-negligible and different contribution towards 
the total prediction variations. 
As we add more randomness sources, the mean prediction of the ensembles 
tends to become more accurate,
and the prediction variations get higher. 
On all three target tasks, the R7 setting appears to exhibit 
the best or close to best accuracy score, and its prediction variations are also 
the highest or close to the highest among all the randomness settings. 
It also seems that different target tasks or datasets 
are sensitive to different types 
of randomness sources. 
We observe that the \criteo\ data is more sensitive to Shuffle 
and Jackknife randomness 
while the \movielens\ data is more sensitive to RandInit. 
We verify that 
by fixing all the randomness sources, we do not observe any prediction variation
in the R0 settings. The PV Mean and PV std is always 0 for R0. 
According to PV coefficient, 
we notice that \movielens\ shows around 5-6\% of 
prediction sway from the mean prediction 
while \criteo\ has around 20\% of prediction sway.  

\vspace{0.1cm}
\noindent {\bf Randomness Source Correlations} ---
Under each randomness setting, we are able to obtain the 
prediction variation for each example. 
Figure~\ref{fig:randomness_correlation} shows
the Pearson correlation of the prediction variations of all 
the examples between each pair of the randomness settings.
The randomness setting can be found at Table~\ref{tab:randomness_control}. 
We do not consider R0 because it eliminates all the randomness 
in the model and PV is always 0.

As we can see from Figures~\ref{fig:randomness_correlation}, 
the prediction variation correlation patterns on \movielens\ is quite 
different from \criteo. 
For example, while the lowest Pearson correlation score on \movielens\ is around 0.7, 
all the randomness settings on \criteo\ are quite correlated with the 
lowest Pearson correlation score to be around 0.92.

\vspace{0.1cm}

\noindent {\bf Regression vs Classification} ---
On the \movielens\ dataset, we are able to predict ratings 
as regression or classification. 
Table~\ref{tab:randomness_control} shows that 
the prediction accuracy is higher when we 
predict ratings as a classification task than as a regression task
almost for all the randomness settings, 
as classification optimizes for the accuracy metric directly.

\begin{figure}
    \centering
    \includegraphics[width=.85\linewidth]{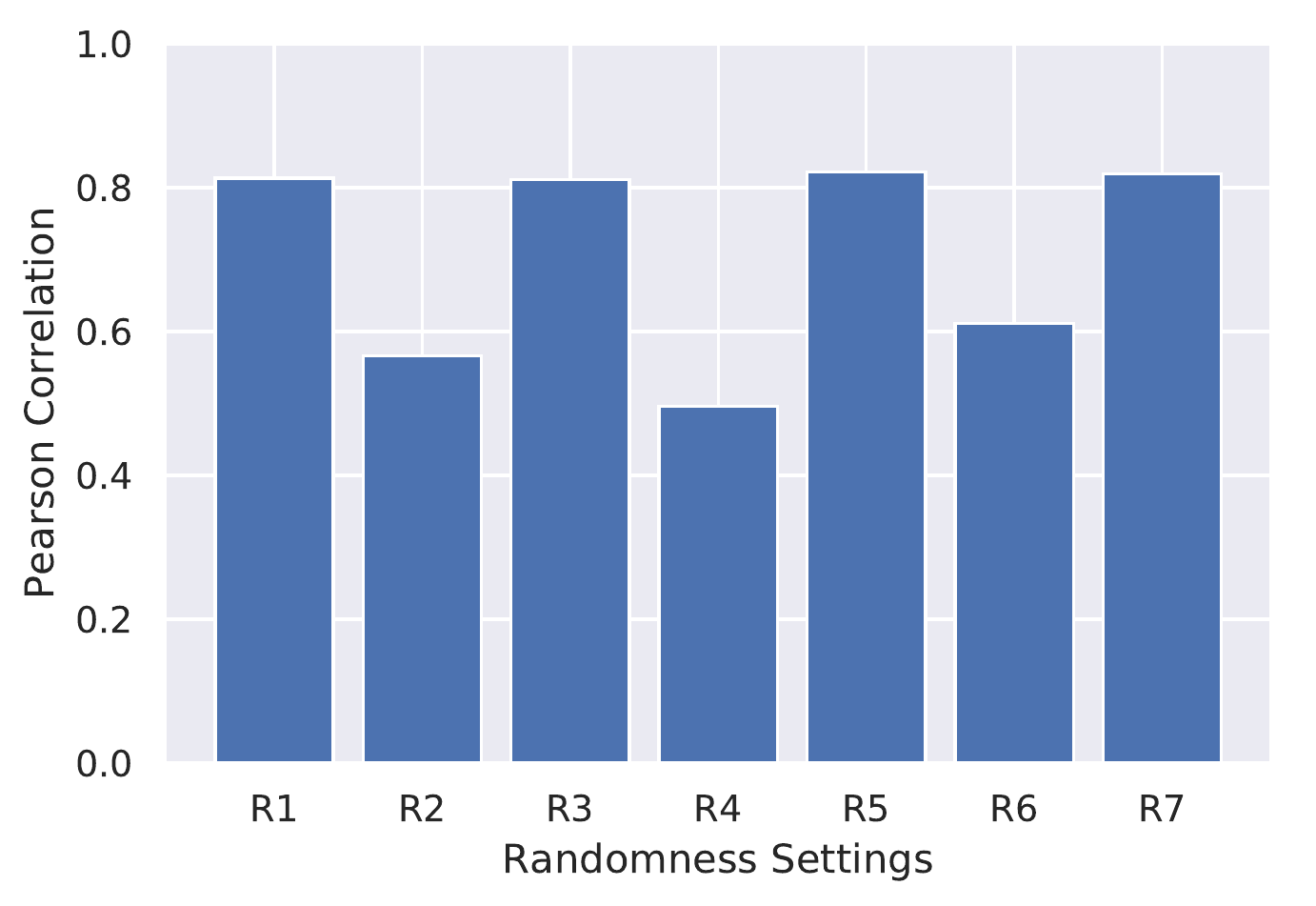}
    \vspace{-0.4cm}
    \caption{Pearson correlation of prediction variations between \mr\ and \mc.}
    \label{fig:correlation_movielens_tasks}
\end{figure}

We also compare the prediction variations obtained 
through regression and classification, and 
Figure~\ref{fig:correlation_movielens_tasks} shows the Pearson correlation 
of prediction variations 
for the two tasks on various randomness settings for all the testing 
examples. We find that whether the variations are strongly 
correlated depends on the randomness settings. 
As shown in the figure, the prediction variations 
are highly correlated (with Pearson correlation more than 0.8) when 
we add the RandInit randomness source, otherwise the two tasks are 
less correlated. 
we think the reason is that RandInit controls model parameters and the loss function 
plays an important role, 
while underlying data properties affect Shuffle and Jackknife more.

\section{Prediction Variation Estimation}
\label{sec:variation_estimation}

In this section, we study the problem of using neuron activation strength to 
infer prediction variation. We first discuss the prediction variation estimation 
task setup in Section~\ref{sec:variation_estimation_setup}, and then show
the experiment results on using neuron activation strength to estimate 
prediction variation in Section~\ref{sec:variation_estimation_results}.

\subsection{Variation Estimation Task Setup}
\label{sec:variation_estimation_setup}

As shown in Figure~\ref{fig:framework}, 
the variation estimation task takes in the neuron activation information 
collected during the target task inference time. 
We use the prediction variation estimated by the ensemble model for training, 
and then during inference time, the variation estimation model is able 
to infer the prediction variation as a cheap auxiliary task. 
We set up the variation estimation task as follow. 

\vspace{0.1cm}
\noindent {\bf Ground-truth Labels} --- 
We first obtain the prediction variation ground-truth from the ensemble. 
On both \movielens\ or \criteo, 
we first split the data into training $D_{t}$ 
and testing $D_{e}$ for the target task.
On \movielens, we split the 1M data 
into 60\% for training and 40\% for testing; 
on \criteo, we use the same setting as in~\cite{ovadia2019can}, 
37M data for training and 4.4M for testing.
We train an ensemble of 100 models to 
obtain $PV(x)$ for each $x \in D_{e}$. 

\vspace{0.1cm}
\noindent {\bf Evaluation Procedure} --- 
We further split $D_e$ into 2 sets: 50\% as 
$D_{e1} = (x_{e1}, PV(x_{e1}))$ for training the 
variation estimation model; and another 50\% as 
$D_{e2} = (x_{e2}, PV(x_{e2}))$ for testing.  

Given a trained target task model $m_t$, 
we build a neural network model $m_v$ to estimate prediction variations. $m_v$ collects the activation 
strength information from $m_t$'s neurons 
during $m_t$'s inference time. 
$m_v$ trains on $D_{e1}$ and tests on $D_{e2}$ with fully connected layers of size $[100, 50]$, 
batch size 256, Adam optimizer 
with learning rate 0.001, and 150 training epochs with early stopping. 
We find that $m_v$ takes less than one epoch to converge on \criteo, 
but takes longer to converge on \movielens\ due to its 
much smaller data size.

Now we explain $m_v$'s input features and objective in detail.

\vspace{0.1cm}
\noindent {\bf Input Features} --- 
We consider two types of input features 
collected from neuron activation strength. 
We use ReLU~\cite{nair2010rectified} as the activation function.
We believe our activation strength feature is general and 
can be applied to other activation functions, such as 
Softplus~\cite{glorot2011deep}, ELU~\cite{clevert2015fast}, 
GELUs~\cite{hendrycks2016gaussian}, 
and Swish~\cite{ramachandran2017searching}.
Due to space limitations, we only experiment with ReLU.

{\em Binary} --- On ReLU neurons, we consider whether a neuron is activated as 
    the input feature. This binary feature represents whether 
    the neuron output is greater than 0. 
    
{\em Value} --- The raw value of a neuron's output directly 
    represents the strength of an activated neuron. Therefore, we experiment with normalized activation value as the input feature. 
    We normalize the neuron outputs according to the neuron output 
    mean and standard deviation collected from the training data.

\vspace{0.1cm}
\noindent {\bf Objective} ---
We estimate the prediction variation in two ways. 

{\em Regression} --- 
    In this setting, the model directly estimates the prediction variation 
    as a regression task. We use Mean Squared Error (MSE) 
    as the loss function. However, by directly optimizing for MSE, 
    this regression task's output range could be huge. As a result, we limit the 
    minimum output of the model to be 0 as 
    prediction variation is always positive, 
    and limit the maximum output to be $mean + 3*std$ where the mean 
    and std are estimated on the training data's prediction variations. 
    $mean + 3*std$ should be able to cover 99.7\% of the data in
    a Guassian distribution.\footnote{
    \url{https://en.wikipedia.org/wiki/68-95-99.7_rule}} 

{\em Classification} --- 
    In this setting, we divide prediction variation 
    into multiple buckets according to the percentile, 
    and then predict which variation bucket it belongs to. 
    We set the bucket number to be 5, and use cross entropy as 
    the loss function for the prediction variation classification model.

\begin{table}[t!]
\begin{tabular}{l|cc|cc|cc}
\toprule
    & \multicolumn{2}{c|}{MovieLens-R} & \multicolumn{2}{c|}{MovieLens-C} &  \multicolumn{2}{c}{Criteo} \\ \cline{2-7}
    
     & MSE  & $R^2$ & MSE  & $R^2$ & MSE  & $R^2$ \\ \hline
(R1) R
& 0.0022 & 0.5416   & 3.6159 & 0.4586
& 0.0063 & 0.7617
\\ 
(R2) S
& 0.0011 & 0.5636  & 1.5015 & 0.4683
& 0.0068 & 0.7863
\\ 
(R3) R+S
& 0.0019 & 0.5514 & 3.4288 & 0.4569
& 0.0062 & 0.8100
\\  \hline 
(R4) J
& 0.0025 & 0.3885 & 2.5986 & 0.3386 
& 0.0086 & 0.7817
\\ 
(R5) R+J
& 0.0024 & 0.5219 & 3.7727 & 0.4646  
& 0.0085 & 0.7868
\\ 
(R6) S+J
& 0.0017 & 0.4938 & 2.3175 & 0.4125
& 0.0091 & 0.7739
\\ 
(R7) R+S+J
& 0.0022 & 0.5123  & 4.0496 & 0.4368
& 0.0092 & 0.7761
\\ \hline
Average
&0.0020	&0.5104	&3.0407	&0.4338	
&0.0078	&0.7824 \\
\bottomrule
\end{tabular}
\caption{Performance of variation estimation as regression
on 7 randomness settings. 
}
\label{tab:uncertainty_prediction_regression}
\end{table}

\subsection{Variation Estimation Performance}
\label{sec:variation_estimation_results}

In this section, we show the variation estimation performance using 
neuron activation strength on \movielens\ and \criteo.

\vspace{0.1cm}
\noindent {\bf Regression Performance} --- 
When we run the prediction variation estimation as a regression task, 
we directly output a score as the estimated prediction variation. 
We use both binary and value input features. 

In Table~\ref{tab:uncertainty_prediction_regression}, 
we show the Mean Squared Error (MSE) and $R^2$ for 
the three target tasks on the 7 randomness control settings.  
From the table, 
on all the three target tasks and all the 7 randomness control 
settings, we observe strong prediction power of neuron activation strength for ensemble prediction variations. 
The average $R^2$ on \mr\ is 0.51, on \mc\ is 0.43, 
and on \criteo\ is 0.78. 
The variation estimation performance is the best 
on the \criteo\ data, 
while \mr\ is better than \mc. The reasons could be: 
First, \criteo\ has more training data than \movielens. 
To train the variation estimation model, 
we have 2.2M (50\% of 4.4M) training data on \criteo, 
while only 0.2M (50\% of 0.4M) training data on \movielens;
Second, \criteo\ has a larger relative range of prediction variations, compared to \movielens.
As shown in Table~\ref{tab:randomness_control}, 
\criteo\ shows around 20\% of prediction variation sway from the mean prediction, which is much higher than ~4-6\% on \movielens; 
Finally, the \criteo\ task is probably the easiest task among 
the three: it is a binary classification task, 
while \mr\ is a regression task and \mc\ is a multi-class
classification task.

\begin{figure}
    \centering
    \includegraphics[width=\linewidth]{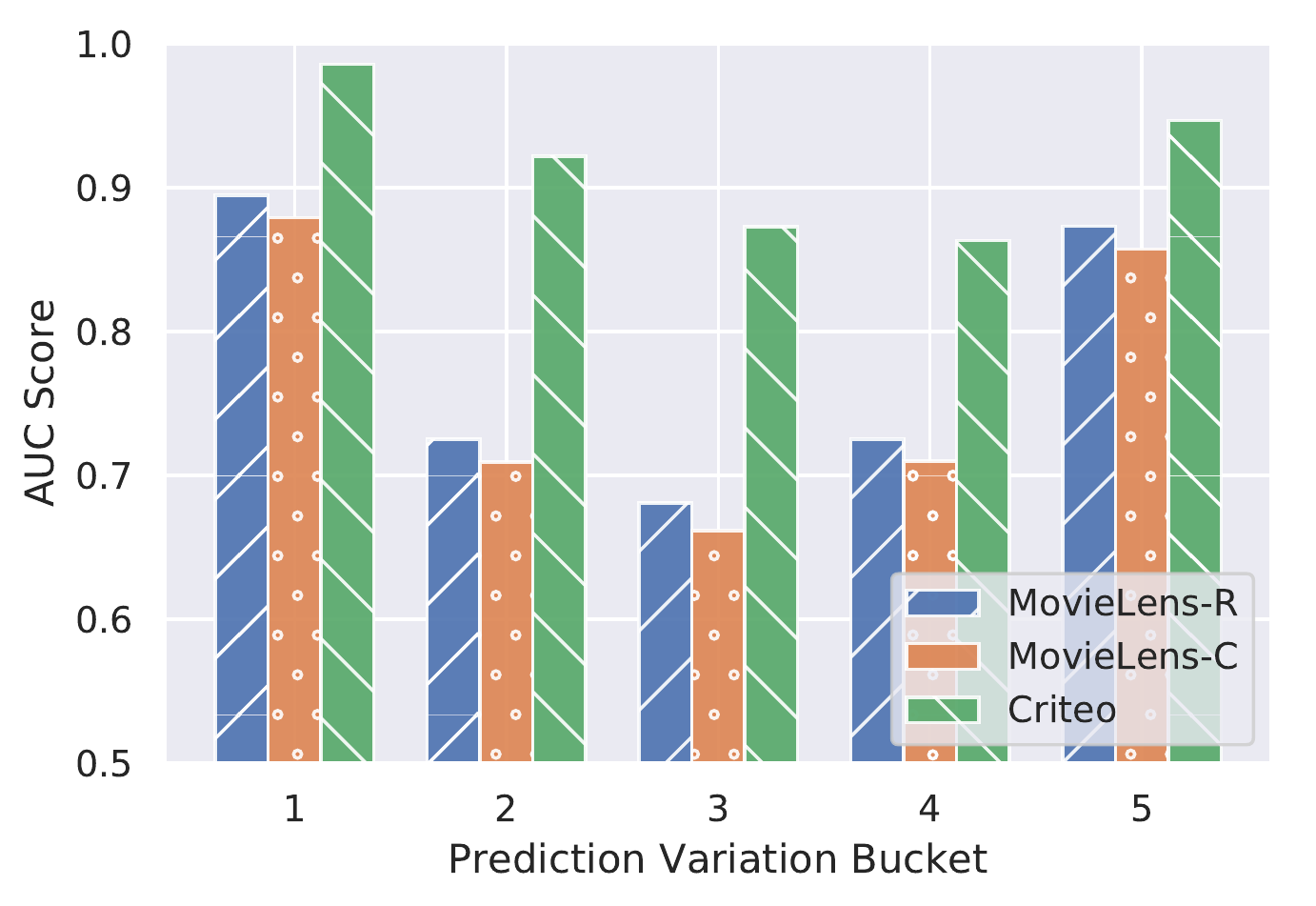}
    \caption{AUC for variation bucket prediction 
    with the randomness setting R3.}
    \label{fig:classification_auc}
\end{figure}

\vspace{0.1cm}
\noindent {\bf Classification Performance} --- 
We also run the prediction variation estimation as a classification task,
by predicting which variation bucket it should be in. 
We use both binary and value input features. Due to the space limit, we 
only show the results on the R3 randomness setting, as 
R3 uses training data shuffling and parameter random initialization which is the most common setting in practice.

\begin{figure*}
    \includegraphics[width=.33\linewidth]{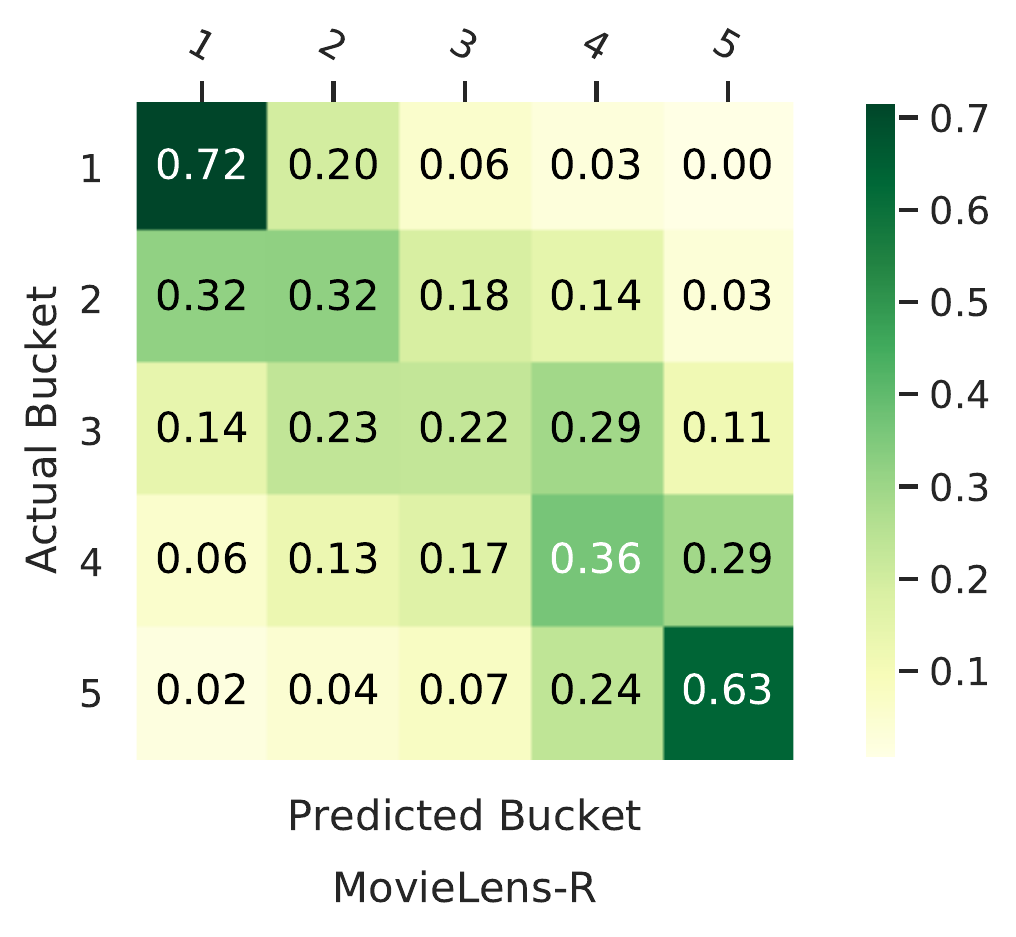}
    \includegraphics[width=.33\linewidth]{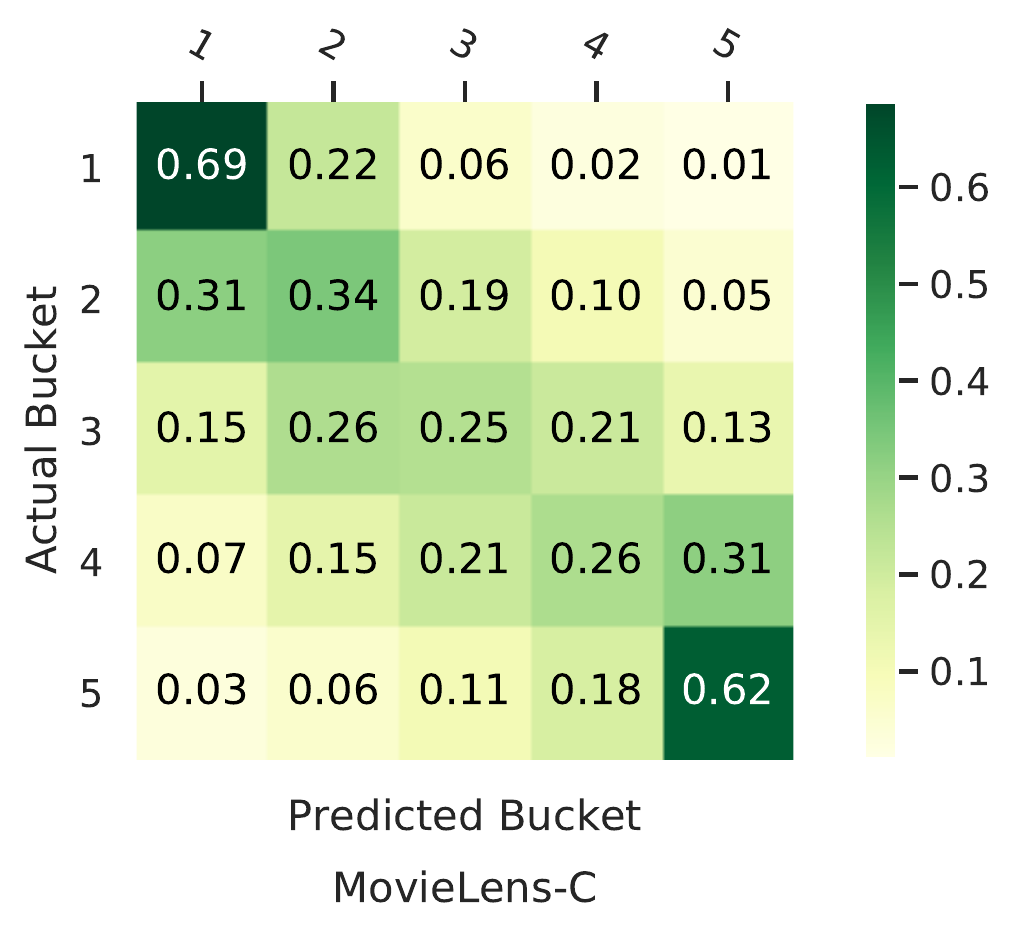}
    \includegraphics[width=.33\linewidth]{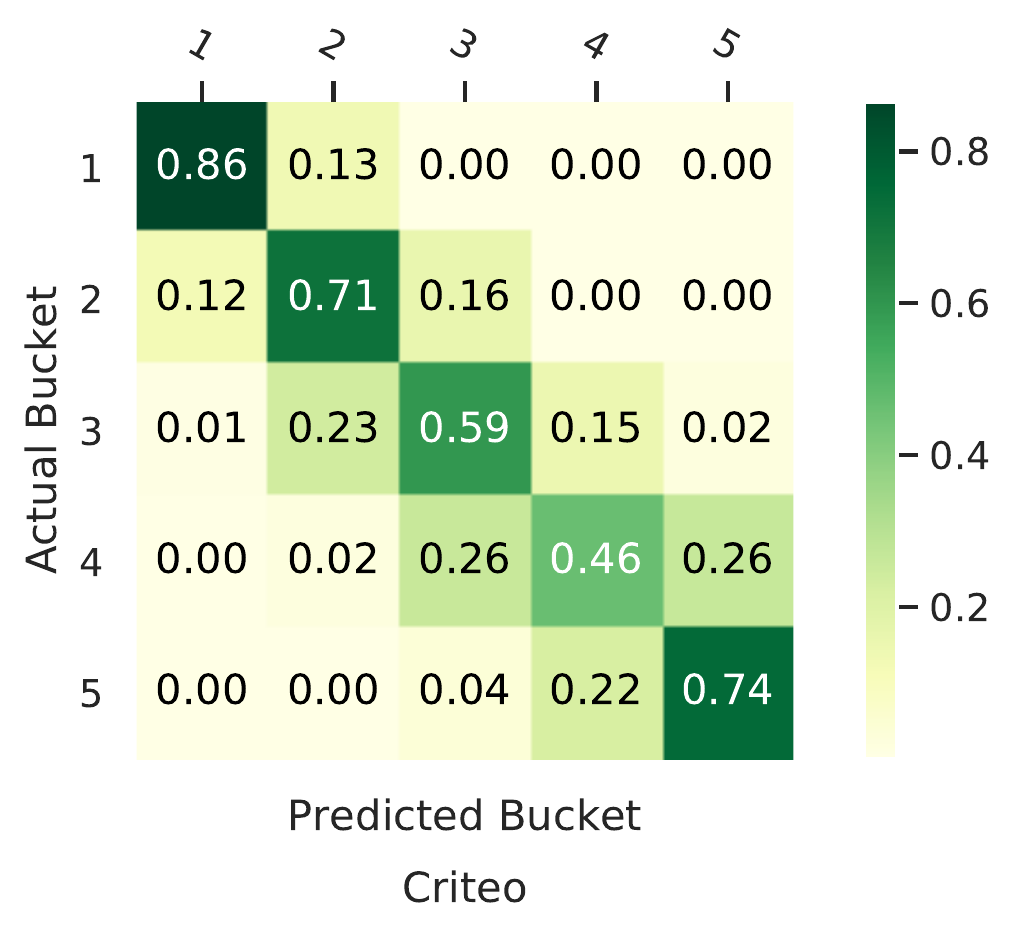}
    \vspace{-0.8cm}
    \caption{Confusion matrix for variation estimation as classification for the randomness setting R3.}
    \label{fig:confusion_matrix}
\end{figure*}

Figure~\ref{fig:classification_auc} shows the AUC scores and
Figure~\ref{fig:confusion_matrix} shows the confusion matrix 
for the 5-bucket prediction variation classification 
on both \movielens\ and \criteo. 
The numbers in Figure~\ref{fig:confusion_matrix} 
are normalized by the actual example number in each bucket. 
Bucket 1 represents the lowest variation slice, while 
bucket 5 represents the highest. 

As we can see from Figure~\ref{fig:classification_auc}, 
our variation estimation model is fairly good at 
distinguishing examples at different variation buckets, especially 
for the lowest and highest buckets. 
The average AUC score for the three tasks on bucket 1 is about 0.92 
and on bucket 5 is about 0.89. 
Figure~\ref{fig:confusion_matrix} shows 
that the classification errors mostly 
happen on adjacent buckets. 
For example, on \criteo, most of the mis-classifications assign bucket 1 examples to bucket 2.
When we divide the prediction variation buckets on training data, 
we notice that the bucket thresholds are close. 
For example, under the randomness control setting R3,
the thresholds of the 5 buckets 
for \mr\ are [0.1420, 0.1672, 0.1950, 0.2366], 
and the thresholds for \criteo\ is [0.0194, 0.0287, 0.0398, 0.0515]. 
As show in the figures, 
\criteo\ seems to have the best performance among the three tasks. 
Again the reasons could be that the \criteo\ task 
has more training data, is probably the easiest 
among the three tasks, 
and it has much larger relative prediction variation range.

\begin{table}[t!]
\begin{tabular}{c|cc|cc|cc}
\toprule
 & \multicolumn{2}{c|}{MovieLens-R} & \multicolumn{2}{c|}{MovieLens-C} &  \multicolumn{2}{c}{Criteo} \\ \cline{2-7}
    
     & MSE  & $R^2$ & MSE  & $R^2$ & MSE  & $R^2$ \\ \hline
B & 0.0025 & 0.4196 & 4.1616 & 0.3408
& 0.0124 & 0.6210
\\
BV & 0.0019 & 0.5514 & 3.4288 & 0.4569 
& 0.0062 & 0.8100
\\
\bottomrule
\end{tabular}
\caption{Activation feature study for variation estimation as 
regression with the randomness setting R3.}
\label{tab:feature_ablation}
\vspace{-0.4cm}
\end{table}

\begin{table}[t!]
\begin{tabular}{c|cc|cc|cc}
\toprule
 & \multicolumn{2}{c|}{MovieLens-R} & \multicolumn{2}{c|}{MovieLens-C} &  \multicolumn{2}{c}{Criteo} \\ \cline{2-7}
    
Run     & MSE  & $R^2$ & MSE  & $R^2$ & MSE  & $R^2$ \\ \hline
1 & 0.0019 & 0.5514 & 3.4288 & 0.4569 
& 0.0062 & 0.8100 \\
2 & 0.0020 & 0.5160 & 3.3243 & 0.4734
& 0.0061 & 0.8140 \\
3 & 0.0019 & 0.5418 & 3.1825 & 0.4959  
& 0.0064 & 0.8036 \\
4 & 0.0019 & 0.5384 & 3.2375 & 0.4872 
& 0.0070 & 0.7863 \\
5 & 0.0021 & 0.4994 & 3.2631 & 0.4831 
& 0.0066 & 0.7969 \\ \hline
Std & 0.00008 & 0.0190 & 0.0842 & 0.0133
&0.0003 &0.0098 \\ 
\bottomrule
\end{tabular}
\caption{Reproducibility test for variation estimation 
as regression with the randomness setting R3.}
\vspace{-0.4cm}
\label{tab:reproducibility}
\end{table}

\vspace{0.1cm}
\noindent {\bf Activation Feature Study} --- 
In Table~\ref{tab:feature_ablation}, we show the contribution of 
the two input activation strength features. 
We try two feature settings: {\em B} refers to using the binary feature only; and 
{\em BV} refers to using both the binary and value features. 
As shown in the table, each type
of features makes a non-negligible contribution towards 
prediction variation estimation. 
Thus, it is beneficial to have both the binary and value features 
for prediction variation estimation. 

\vspace{0.1cm}
\noindent {\bf Reproducibility} ---
When we evaluate the variation estimation model, the performance is calculated 
based on one target task model. 
To check whether the performance is reproducible, 
we train 5 new target task models. 
To simplify the problem, we also use the randomness setting R3, which 
is the most commonly used setting in practice. 
Table~\ref{tab:reproducibility} shows the MSE and $R^2$ for each run on 
both \movielens\ and \criteo. As shown in the table, 
the standard deviation of MSE and $R^2$ for the 5 runs is small for 
each of the three tasks. 
Thus, we conclude that activation strength is useful for estimating 
prediction variation and it is reproducible.

\vspace{0.1cm}
\noindent {\bf Comparison with Dropout~\cite{gal2016dropout}} --- 
Dropout is also a standard way to estimate model uncertainty. 
We estimate prediction variation using dropout as follows: 
We train one model by randomly dropping 20\% of the neurons on the ReLU 
layers using the randomness settings R3. During inference time, we
keep the dropout turned on to obtain the predicted results 
for all the testing data. 
We run the inference 100 times, and 
obtained the prediction variation for each testing example.  
We find that the prediction variation estimated by dropout is not very correlated 
with the variation estimated by the ensemble method. 
On \mr, Pearson correlation of prediction variations for dropout
and ensemble is  0.25, RMSE is 0.0798, and $R^2$ is  -0.5010. 
On \criteo, Pearson correlation of prediction variations for dropout
and ensemble is 0.37, RMSE is 0.0279, and $R^2$ is -1.3709. 
As a result, we did not conduct further comparison with 
our activation strength based method.

\section{Conclusion and Future Work}
\label{sec:conclusion}

In this paper, we conduct empirical studies to 
understand the prediction variation estimated by 
ensembles under various randomness control settings.  
Our experiments on two public datasets 
(\movielens\ and \criteo) demonstrate that 
with more variation sources, ensemble tends to 
produce more accurate point estimates 
with higher prediction variations.
More importantly, we demonstrate strong predictive power of neuron activation strength to infer ensemble prediction variations, which provides an efficient way to estimate prediction variation without the need to run inference multiple times as in ensemble methods.

In the future, we are interested in exploring the proposed activation 
strength based methods in various applications, such as 
model-based reinforcement learning, and curriculum learning.
In addition, we plan to make two further improvements to our 
activation strength based approach. 
First, we would like to study additional neuron activation patterns, 
such as adjacent neuron paths, to further
improve the variation estimation model. 
Second, currently we re-train the variation 
estimation model for each new target task. 
Activation pattern is a universal and general feature, 
and we hope to find 
a universal model for prediction variation estimation 
without re-training each individual target task.

\appendix

\section{Appendix}

\begin{figure}[t!]
    \centering
    \includegraphics[width=\linewidth]{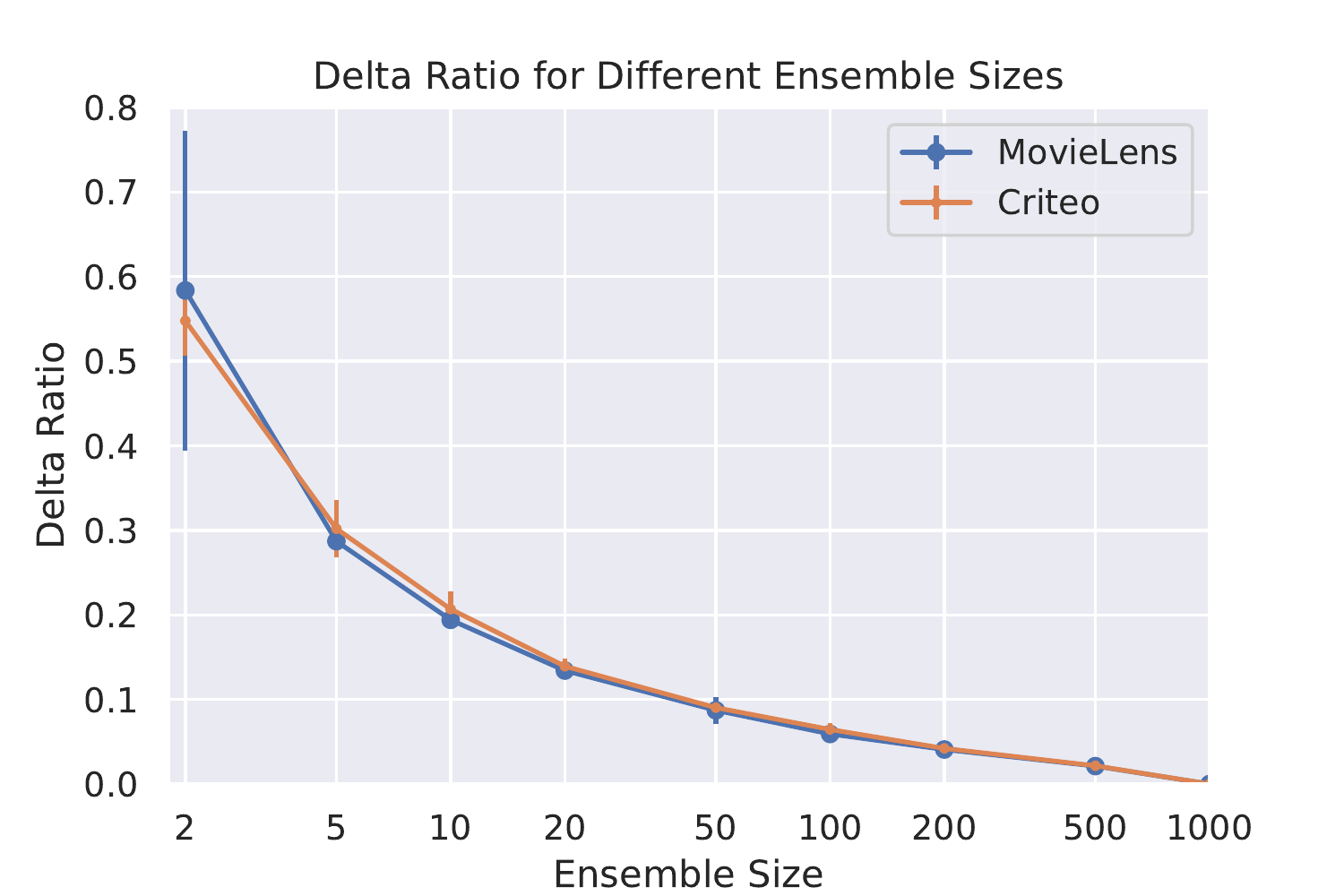}
    \caption{Delta ratio for different ensemble sizes on 
    the MovieLens regression task and the \criteo\ task.}
    \vspace{-0.4cm}
    \label{fig:ensemble_size}
\end{figure}

\subsection{Model Ensemble Sizes}
\label{sec:ensemble_size}

Our prediction variation is estimated using the ensemble method, and 
we are interested in finding out how many models to ensemble to estimate prediction variation accurately. In this section, we 
conduct empirical experiments on \mr, 
and \criteo\ to answer this question. 

For each target task, using the same training data and model configuration, 
we first train 1000 models as the ensemble universe $M_{gt}$ 
to obtain ground-truth prediction variations.
We used the R3 randomness setting here, as it is the most commonly 
used setting in practice. 
Given an example $x$, we obtain its 
prediction variation from the 1000 ensemble models as $PV_{gt}(x)$ . 
We calculate the mean prediction variation for all the examples 
as $\bar{PV}_{gt}$.

Then we evaluate the prediction variation difference between an ensemble $M_N$ of a smaller size $N$ and 
$M_{gt}$. 
We use {\em delta ratio} to quantify the difference 
between the prediction variation estimated from the 
two ensembles $M_N$ and $M_{gt}$ as follows. 

\begin{definition} (Delta Ratio)
Let prediction variation delta 
$\delta_{M_N}(x)$ be the absolute difference of the estimated prediction variation
between a model ensemble $M_N$ of size $N$ and 
the ground-truth model ensemble $M_{gt}$, as 
$\delta_{M_N}(x) = |PV_{M_N}(x) - PV_{gt}(x)|$. 
We obtain the average prediction variation delta for all the examples 
in a dataset $D$ as 
$\delta_{M_N} =\frac{1}{|D|}\sum_{x \in D}\delta_{M_N}(x)$. 
We define delta ratio $dr_{M_N}$ 
to be the ratio of prediction variation delta $\delta_{M_N}$
to the average prediction variation in the ground-truth ensemble models $\bar{PV}_{gt}$, as 
$dr_{M_N} = \delta_{M_N} / \bar{PV}_{gt}$ 
\end{definition}

In Figure~\ref{fig:ensemble_size}, 
we show the delta ratio of different ensemble sizes for 
the \movielens\ regression task and the \criteo\ task. 
For each ensemble size $N$, we sample N models without replacement
from the 1000 ground-truth model universe and obtain its delta ratio. 
We repeat this sampling process for 20 times, and 
obtain the mean and standard deviation of the delta ratio for the given $N$. 
We plot the delta ratio as shown in Figure~\ref{fig:ensemble_size}. 
We can see the delta ratio decreases when the ensemble size increases. 
The delta ratio statistics is similar on both datasets, \movielens\ and \criteo.
When the ensemble size is 100, the delta ratio is about 7\% which 
indicates 93\% of prediction variation from the ground-truth ensemble 
of 1000 models is captured. 
As a result, in this paper, we use 100 as the default 
ensemble size for all experiments.

\bibliographystyle{ACM-Reference-Format}
\bibliography{reference}

\end{document}